\pdfoutput=1

\documentclass[11pt]{article}

\usepackage{acl}

\usepackage{times}
\usepackage{latexsym}

\usepackage[T1]{fontenc}

\usepackage[utf8]{inputenc}

\usepackage{microtype}

\usepackage{graphicx}
\usepackage{tabularx}
\usepackage{cleveref}
\usepackage{listings}
\usepackage{subcaption}
\usepackage{arydshln}

\title{Evaluating the Text-to-SQL Capabilities of Large Language Models}

\author{Nitarshan Rajkumar$^1$\thanks{\hspace{1.5mm}Work partially done at Mila and the Université de Montréal.}\phantom{\footnotesize 1}, Raymond Li$^2$, Dzmitry Bahdanau$^2{}^3{}^4{}^5$\\
  $^1$University of Cambridge, $^2$ServiceNow,
  $^3$Mila,
  $^4$McGill University,
  $^5$Canada CIFAR AI Chair
  \\
  \small{\texttt{\href{mailto:nr500@cam.ac.uk}{nr500@cam.ac.uk}}}, \small{\texttt{\{\href{mailto:raymond.li@servicenow.com}{raymond.li},\href{mailto:dzmitry.bahdanau@servicenow.com}{dzmitry.bahdanau}\}@servicenow.com}}\\
  \small{\url{https://github.com/nitarshan/codex-text2sql}}}

\begin{document}
\maketitle
\begin{abstract}
We perform an empirical evaluation of Text-to-SQL capabilities of the Codex language model.
We find that, \textit{without any finetuning}, Codex is a strong baseline on the Spider benchmark; we also analyze the failure modes of Codex in this setting.
Furthermore, we demonstrate on the GeoQuery and Scholar benchmarks that a small number of in-domain examples provided in the prompt enables Codex to perform better than state-of-the-art models finetuned on such few-shot examples.
\end{abstract}

\section{Introduction}
Translating natural language questions to SQL queries (Text-to-SQL) is an important business problem which has seen significant research interest.
A common approach to this task involves training a model to produce a SQL query when given a question, a database schema, and possibly database content as inputs.
A clear trend in this area is to finetune models pretrained on natural language; notably, performance significantly improves as larger pretrained models are used \citep{shaw-etal-2021-compositional,scholak2021picard}.

Recent results from the broader field demonstrate that simply scaling training data and model size for generative language models brings advanced capabilities, such as few-shot learning without finetuning \citep[GPT-3, ][]{brown2020language} and code generation \citep[Codex, ][]{chen2021evaluating}.
In this work we study if such models are already competitive Text-to-SQL solutions \textit{without any further finetuning on task-specific training data}, evaluating Codex and GPT-3 models of different sizes with varied prompts on Text-to-SQL benchmarks.

We find that Codex achieves a competitive performance of up to 67\% execution accuracy on the Spider development set.
We analyze the predicted queries that automatic evaluation judged as wrong and find that many of them would be judged correct by humans, whereas others could likely be fixed within the no-finetuning paradigm. 
Lastly, using  GeoQuery and Scholar benchmarks we show that adapting Codex to a specific domain by prompting it with few examples can be more effective than fine-tuning a smaller language model on the same examples.

\section{Experimental Setup}

\begin{table}[t]
    \centering
    \small
    \begin{tabular}{lcccc}
        \hline
        \textbf{Model} & \textbf{VA} & \textbf{EX} & \textbf{TS} \\
        \hline
        \multicolumn{4}{l}{\textit{Finetuned}}\\
        \hline
        T5-base & 72.7 & 57.9 & 54.5 \\
        T5-large & 84.1 & 67.2 & 61.4 \\
        T5-3B & 87.6 & 71.4 & 65.7 \\
        T5-3B$^*$ & 88.2 & 74.4 & 68.3 \\
        T5-3B + PICARD$^*$ & 97.8 & 79.1 & 71.7 \\
        BRIDGE v2$^*$ & -- & 68.0 & -- \\

        \hline
        \multicolumn{4}{l}{\textit{Inference-only}}\\
        \hline
        GPT-3 ada & 33.8 & 2.3 & 0.3 \\
        GPT-3 babbage & 48.8 & 5.7 & 3.9 \\
        GPT-3 curie & 70.9 & 12.6 & 8.3 \\
        GPT-3 davinci & 65.0 & 26.3 & 21.7 \\
        Codex cushman$^*$ & 86.3 & 63.7 & 53.0 \\
        Codex davinci$^*$ & 91.6 & 67.0 & 55.1 \\

    \end{tabular}
    \caption{
    Best Spider development set performance across models, as measured by percentage of predictions which are valid SQL (VA), execution accuracy (EX), test-suite accuracy (TS).
    Models marked with $^*$ use database content.
    T5 results are from \citet{scholak2021picard}, BRIDGE v2 results are from \citet{lin2020bridging}.
    }
    \label{tab:engines}
\end{table}

\textbf{Models}
Our evaluation focuses on the models accessible via the OpenAI API: GPT-3 (in the ascending ada, babbage, curie and davinci sizes) and Codex (in the ascending cushman-codex and davinci-codex sizes)\footnote{See \Cref{sec:app:api:paramcount} for a discussion on parameter counts.}.
These are generative language models which perform next-token prediction during training and inference; GPT-3 is trained on a diverse set of sources from the internet, and Codex is further finetuned on code from GitHub.
We compare GPT-3 and Codex against  methods from \citet{shaw-etal-2021-compositional} using the T5 encoder-decoder model.
Starting from public checkpoints pretrained on Common Crawl, the T5 model is finetuned on Spider to predict the output SQL, conditioned on the question and schema.
The 3B parameter T5 model is currently the state-of-the-art on Spider when combined with constrained inference using the PICARD algorithm \citep{scholak2021picard}.
We also compare to BRIDGE v2 \citep{lin2020bridging}, a sequence-to-sequence model based on BERT.

\label{subsec:datasets}
\textbf{Zero-Shot Experiments}
We use the Spider benchmark \citep{yu2019spider} for cross-domain Text-to-SQL.
We report performance using percentage of development set predictions which are valid (executable) SQLite SQL, execution accuracy, and test-suite execution accuracy.
The latter metric was proposed by \citet{zhong2020semantic} to measure semantic equivalence of SQL queries written in different styles, which is essential when comparing Codex to models trained on Spider.
We address concerns around possible memorization of Spider data by Codex in \Cref{sec:app:api:memorization}.

\textbf{Few-Shot Experiments}
We re-purpose the question-splits of the GeoQuery and Scholar datasets \citep{data-geography-original, data-scholar-original, finegan-dollak-etal-2018-improving} to perform experiments in a few-shot setting.
The examples in these datasets are grouped by query templates.
Examples corresponding to the same template have the same SQL query structure, but may have different English questions and SQL literals.
To define the few-shot task, we first sort the templates by their frequency in the training set. In the $n$-shot setting we then use one random example for each of the $n$ most frequent templates.

\textbf{Prompts}
We use six prompt structures in our experiments (examples provided in \Cref{sec:appendix_prompts}).
\textbf{Question} provides no database information and just includes the question as a SQL comment.
\textbf{API Docs} follows the style of the Text-to-SQL example in Codex documentation and includes a schema in a comment style which does not conform to SQLite standards.
\textbf{Select X} includes in comments the results of executing a \texttt{SELECT * FROM T LIMIT X} query on each table, including schemas via column headers.
\textbf{Create Table} includes the \texttt{CREATE TABLE} commands for each table, including column type and foreign key declarations.
\textbf{Create Table + Select X}\footnote{Only the davinci-codex model can evaluate Create Table + Select X prompts with more than 1 row, due to its expanded 4096-token prompt window compared to the 2048-token window of all other models.
In addition, GPT-3 models preprocess whitespace tokens less efficiently than Codex models, and therefore cannot evaluate Create Table + Select X prompts at all.} is a combination of the preceding two prompt formats.
Finally, \textbf{Fewshot} additionally includes question-query pairs.

\begin{table}[t]
    \centering
    \small
    \begin{tabular}{lccc}
        \hline
        \textbf{Prompt} & \textbf{VA} & \textbf{EX} & \textbf{TS} \\
        \hline
        Question & 14.0 & 8.3 & 8.2 \\
        API Docs & 83.8 & 56.8 & 47.5 \\
        Select 1 & 86.3 & 60.9 & 52.0 \\
        Select 3 & 85.8 & 60.3 & 52.2 \\
        Select 5 & 85.2 & 60.5 & 51.5 \\
        Select 10 & 86.0 & 60.8 & 51.2 \\
        Create Table & 89.8 & 59.9 & 50.0 \\
        $+$ Select 1 & 92.5 & 64.8 & 53.7 \\
        $+$ Select 3 & 91.6 & 67.0 & 55.1 \\
        $+$ Select 5 & 91.0 & 65.3 & 53.9 \\
        $+$ Select 10 & 91.2 & 63.3 & 52.4 \\

    \end{tabular}
    \caption{
    Spider development set performance across prompt styles on the davinci-codex model, as measured by percentage of predictions which are valid SQL (VA), execution accuracy (EX), test-suite accuracy (TS).
    }
    \label{tab:prompts}
\end{table}

\section{Zero-Shot Results}

\begin{figure*}[t!]
    \centering
    \small
    \setlength\tabcolsep{4pt}
    \begin{tabularx}{\linewidth}{l X}
        14\% & \textbf{Shortcuts}\\
        \textbf{NL} & \textit{What is the number of car models created by the car maker American Motor Company?}\\
        \textbf{Gold} & \texttt{SELECT count(*) FROM CAR\_MAKERS AS T1 JOIN MODEL\_LIST AS T2 ON T1.Id  =  T2.Maker WHERE T1.FullName  =  'American Motor Company';}\\
        \textbf{Pred} & \texttt{SELECT COUNT(Model) FROM model\_list WHERE Maker = 1;}\\
        \hdashline
        \textbf{NL} & \textit{Give the city and country for the Alton airport.} \\
        \textbf{Gold} & \texttt{SELECT City ,  Country FROM AIRPORTS WHERE AirportName  =  "Alton"} \\
        \textbf{Pred} & \texttt{SELECT City, Country FROM airports WHERE AirportCode = 'ALN';} \\
        \hline

        8\% & \textbf{SELECT Extra Columns}\\
        \textbf{NL} & \textit{List names of conductors in descending order of years of work.}\\
        \textbf{Gold} & \texttt{SELECT Name FROM conductor ORDER BY Year\_of\_Work DESC}\\
        \textbf{Pred} & \texttt{SELECT Name, Year\_of\_Work FROM conductor ORDER BY Year\_of\_Work DESC;}\\
        \hline
        
        5\% & \textbf{SELECT Convention}\\
        \textbf{NL} & \textit{What are all the makers and models?} \\
        \textbf{Gold} & \texttt{SELECT Maker ,  Model FROM MODEL\_LIST;} \\
        \textbf{Pred} & \texttt{SELECT DISTINCT car\_makers.Maker, model\_list.Model FROM car\_makers JOIN model\_list ON car\_makers.Id = model\_list.Maker;} \\
    \end{tabularx}
    \caption{
    Examples of error types, as made by the davinci-codex model with Create Table + Select 3 prompt.
    NL stands for natural language question.
    Percentage indicates the percentage of errors which are of the given type.
    Further examples are provided in \Cref{fig:extra_error_examples} in \Cref{sec:app_addition}.
    }
    \label{fig:error_examples}
\end{figure*}

We present results for different  model sizes in \Cref{tab:engines} and for different prompt styles in \Cref{tab:prompts}.
Full results are available in \Cref{tab:full_results} in \Cref{sec:app_addition}.

\textbf{Codex provides a strong baseline for Text-to-SQL tasks}
In \Cref{tab:engines} the best performing model (davinci-codex, Create Table + Select 3) achieves 67\% execution accuracy and 56.5\% test suite execution accuracy on Spider.
This is comparable to the performance of the BRIDGE v2 \citep{lin2020bridging} model which achieved a (then) state-of-the-art 68\% execution accuracy in December 2020.

\textbf{Prompt design is critical for performance}
As seen in \Cref{tab:prompts}, providing the question alone results in a low 8.3\% execution accuracy.
There is a progressive improvement to 56.8\% as schema information is introduced in API Docs, to 59.9\% when valid SQL and foreign key information is used in Create Table, and to 67.0\% when database content is introduced with Create Table + Select 3.

\textbf{More database content can harm performance}
In \Cref{tab:prompts} we observe that for the Select Limit X prompts there is a negligible change in performance when adding more rows.
By contrast, Create Table + Select Limit X prompt accuracy peaks with 3 rows before significantly decreasing in performance as more rows are added.

\textbf{Diminishing returns for Codex model size}
While GPT-3 performance significantly benefits from increased model size, the davinci-codex model does not perform drastically better than cushman-codex.
Full results in \Cref{tab:full_results} in \Cref{sec:app_addition} show cushman-codex generally being within 1 percentage point of davinci-codex for the same prompt style; it even performs 3 percentage points \textit{better} for the Create Table prompt.
These results suggest that davinci-codex's longer context window may be a greater contributor to its peak performance than increased parameter count.

\subsection{Error Analysis}
\label{sec:error_analysis}

We focus our error analysis on the davinci-codex model with Create Table + Select 3 prompt, and present a breakdown of prediction types in \Cref{tab:error_percentages} and examples of errors in \Cref{fig:error_examples}.
Our error categories were chosen to surface the most interesting Codex-specific behaviours we observed amongst the errors made.
We randomly selected and annotated 100 predictions which were valid SQL yet were judged incorrect by test-suite evaluation.

We first consider \textbf{Semantic Incorrect} behaviours, which Spider evaluation and the human annotator both view as incorrect predictions.
\textbf{Shortcut} errors are where Codex made use of either specific table values or ``world knowledge'' from GPT-3 pretraining, while the ground-truth query contained the exact literals from the question. 
\textbf{GROUP BY Convention} errors are where Codex incorrectly groups on a non-primary-key column (such as a name or title column).

We also consider \textbf{Ambiguous Correct} behaviours which are semantically different from the gold query and are therefore judged as incorrect by Spider evaluation, but which the human annotator viewed as being an acceptable SQL translation of the given question.
\textbf{SELECT Convention} errors are where Codex selects a different column than the per-database convention of the gold queries (such as name instead of ID).
\textbf{SELECT Extra Columns} errors are where Codex includes additional useful columns in its query beyond what the gold query includes.
\textbf{Argmax} errors are where Codex differs from the gold query in how a min/max resolution (such as ``youngest singer'') is handled for ties.

\begin{table}[t]
    \centering
    \small
    \begin{tabular}{lll}
        \hline
        \textbf{Annotation} & \textbf{\%} & \textbf{E\%} \\
        \hline
        Test-Suite Correct & 55.1 & -- \\
        \hline
        Semantic Incorrect & 25.2 & 69 \\
        -- Shortcuts & 5.1  & 14 \\
        -- \texttt{GROUP BY} Convention & 1.5 & 4 \\
        -- Other & 18.6 & 51 \\
        \hdashline
        Ambiguous Correct & 11.3 & 31 \\
        -- \texttt{SELECT} Extra Columns & 2.9 & 8 \\
        -- \texttt{SELECT} Convention & 1.8 & 5 \\
        -- Argmax & 1.5 & 4 \\
        -- Other & 5.1 & 14 \\
        \hline
        Invalid SQL & 8.4 & -- \\
        -- Ambiguous column name & 1.9 & -- \\
        -- No such column & 4.5 & -- \\
    \end{tabular}
    \caption{
    Breakdown of prediction annotations over Spider development set for the davinci-codex model with Create Table + Select 3 prompt.
    \% is percentage of all predictions, E\% is percentage of manually annotated erroneous queries (see Section \Cref{sec:error_analysis} for details).}
    \label{tab:error_percentages}
\end{table}

We observe in \Cref{tab:error_percentages} that a significant 31\% of valid yet erroneous predictions are penalized by Spider evaluation as being incorrect though a human annotator viewed them as acceptable solutions.
Future work could be to investigate to what extent one can control the behaviour of Codex. This could allow to fix these ambiguous errors, either by prompt design or using a few examples.

\section{Few-Shot}

We investigate whether Codex can perform few-shot Text-to-SQL.
As described in Section \ref{subsec:datasets}, we re-purpose the GeoQuery and Sholar datasets in a few-shot setting.
It is well known that models trained on Spider transfer poorly to other single-database Text-to-SQL datasets \citep{suhr-etal-2020-exploring} in a zero-shot setting.
Studying few-shot Text-to-SQL on GeoQuery and Scholar should show to what extent models are able to leverage a small amount of examples to effectively adapt to a new domain.

\textbf{Baseline}
The baseline is a T5-3B model that was finetuned on Spider, reaching 71\% exact-match accuracy on Spider validation set.
The model is then further finetuned on the new domain -- GeoQuery or Scholar.
The learning rate for domain-specific-finetuning was selected in the $20$-shot setting among $[0.1, 0.2, 0.5, 1, 2]\cdot10^{-5}$, based on the best validation set performance after 300 steps.
We use batch-size 1024, such that all the few-shot examples fit in the same batch.

\textbf{Codex}
Building on the Create Table + Select X prompt, we append $n$ question-query examples to the input in an $n$-shot setting.
An example of this prompt is provided in \Cref{fig:prompt_fewshot}.
All samples are generated using greedy decoding, with temperature $0$.
Note that for a given $n$-shot setting, the baseline and Codex use the same set of support examples.
These examples are in the prompt for Codex, and used to finetune the baseline on the new domain.
Given the limited window-size of API models, on GeoQuery we can feed up to 40 support examples to davinci-codex, and up to 10 examples to cushman-codex and GPT-3 models.
On Scholar the queries are longer and the schema more complex -- we fit only 10 examples in the prompt of davinci-codex, 5 for cushman-codex, and none at all for GPT-3 models.

\subsection{Results}

\begin{figure}[t]
    \centering
    \begin{subfigure}[b]{\linewidth}
         \centering
        \includegraphics[width=\linewidth]{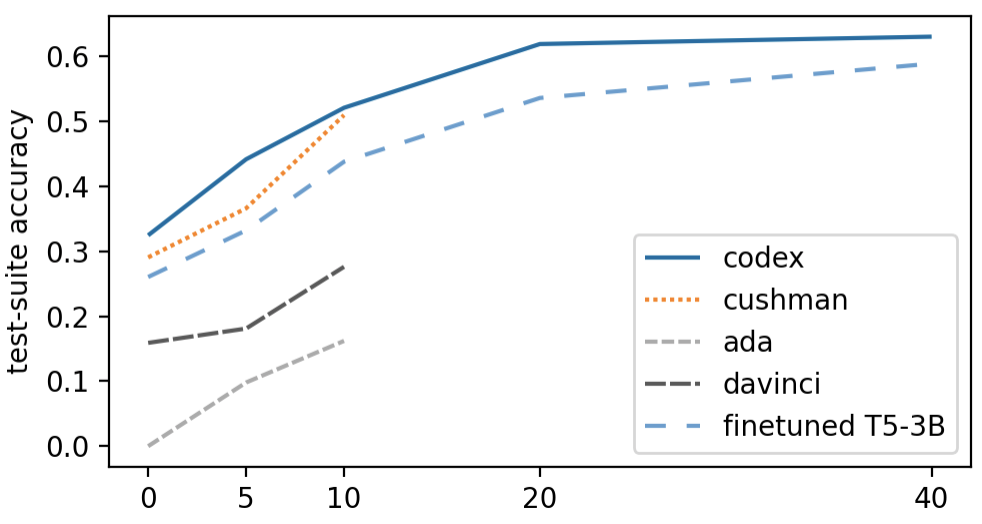}
         \caption{GeoQuery. When trained on the whole GeoQuery training set (549 examples), the finetuned T5 reaches 85.7\% accuracy.}
         \label{fig:few-shot-test}
     \end{subfigure}
    \begin{subfigure}[b]{\linewidth}
         \centering
        \includegraphics[width=\linewidth]{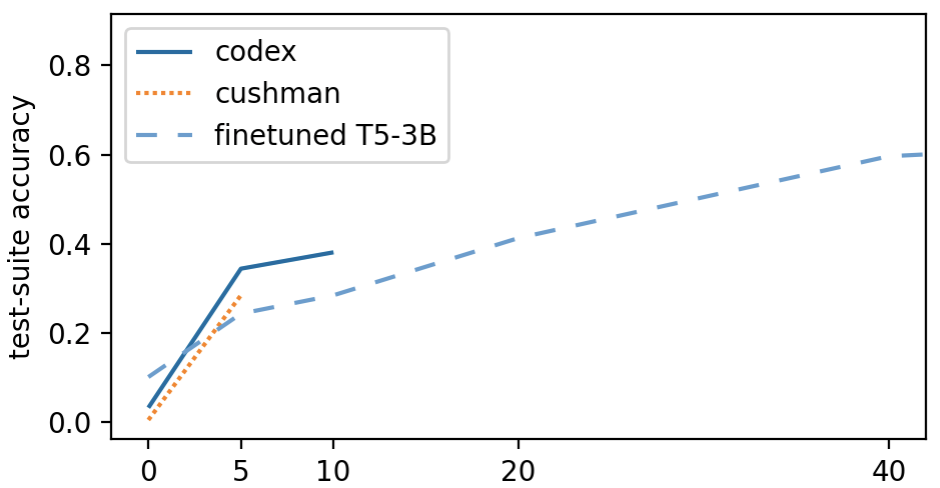}
         \caption{Scholar. When trained on the whole Scholar training set (499 examples), the finetuned T5 reaches 87.2\% accuracy.
         }
         \label{fig:few-shot-test}
     \end{subfigure}
    \caption{
    Test-suite accuracy with varying number of support examples. The x-axis shows the number of few-shot examples used.
    }
    \label{fig:few-shot-num-support}
\end{figure}

Figure~\ref{fig:few-shot-num-support} shows test-suite accuracies on the Scholar and GeoQuery datasets.
The baseline reaches 85.7\% test-set performance when trained on the complete GeoQuery training set (549 examples).
Respectively, it reaches 87.2\% test accuracy when trained on the whole Scholar training set (499 examples).
This simple baseline is a very competitive model when considering the entire datasets.
However Figure~\ref{fig:few-shot-num-support} shows that it is largely beaten by Codex in few-shot settings.
In a zero-shot setting, both davinci-codex and cushman-codex already beat the baseline on GeoQuery.
We speculate that Codex performs well here because it uses the same argmax convention as the GeoQuery dataset, which is different than the convention used in Spider.
With up to 40 examples in the prompt, davinci-codex outperforms a T5-3B model finetuned on these same examples by a large margin, whereas GPT-3 davinci performs quite poorly on this task.
On the other hand, the T5 model outperforms Codex in a zero-shot setting on Scholar. In 5 and 10-shot settings, Codex shows better adaptation from these few samples and beats the T5 baseline.

\section{Conclusion}
We demonstrated that generative language models trained on code provide a strong baseline for Text-to-SQL.
We also provided analysis of failure modes for these models, which we hope guides further prompt design (whether few-shot or through natural language instructions) in this setting.
Finally, we showed that prompt-based few-shot learning with these models performs competitively with finetuning-based few-shot learning of smaller models.
A clear direction for future work is to evaluate the benefits of finetuning with Codex models.

\section*{Acknowledgements}

Nitarshan performed all zero-shot and finetuning experiments as well as error-analysis, and wrote most of the paper.
Raymond performed all few-shot experiments and the associated writing.
Dzmitry supervised, and contributed to paper editing.

We thank Dóra Jámbor for insightful discussions, Laurent Charlin for providing funding for Nitarshan and for providing feedback on this work, Fraser Kelton and Dave Cummings for support with the OpenAI API, and Ruiqi Zhong for assistance with Spider test suites.
We also thank anonymous ARR reviewers for their feedback and criticism in the review process.

Nitarshan additionally thanks the city of Montréal and its cafés for providing inspirational settings in which to conduct this work.

\bibliography{paper}
\bibliographystyle{acl_natbib}

\appendix

\section{API Details}
\label{sec:app:api}

At time of writing, the OpenAI API was accessible at \url{https://openai.com/api/}.
The example from which our API Docs prompt draws from can be found at \url{https://beta.openai.com/examples/default-sql-translate}.

\subsection{Hyperparameters}
\label{sec:app:api:hparam}
We sample 200 tokens from GPT-3 and Codex with temperature 0, with the following strings used as stop tokens to halt generation: ``-{}-'', ``\textbackslash n\textbackslash n'', ``;'', ``\#''.

\subsection{Parameter Counts}
\label{sec:app:api:paramcount}
Parameter counts for OpenAI API models are not openly available.
\citet{gao_sizes_2021} evaluated API GPT-3 models across a variety of language modelling tasks to compare to published results in \citet{brown2020language}, finding that ``Ada, Babbage, Curie and Davinci line up closely with 350M, 1.3B, 6.7B, and 175B respectively''.
We presume that the davinci-codex model is the same size as the GPT-3 davinci model; cushman-codex is a new model name so we can only guess that it is of a similar (but not the same) size to GPT-3 curie.
Nevertheless these remain guesses which should not be relied on.

\subsection{Model Versioning}
\label{sec:app:api:versioning}
The exact models served through the OpenAI API may vary over time.
We verified that for each model type, only a single model version was used to generate results.
These versions are
\texttt{ada:2020-05-03}, \texttt{babbage:2020-05-03}, \texttt{curie:2020-05-03}, \texttt{davinci:2020-05-03}, \texttt{cushman-codex:2021-08-03}, \texttt{davinci-codex:2021-08-03}.

\subsection{Finetuning}
In \Cref{tab:full_results} we include preliminary results from finetuning GPT-3 models on the Spider training set.
We used the full training set, and the default finetuning settings of 4 epochs, a batch size of 8, and a learning rate multiplier of 0.1.
We did not perform a hyperparameter sweep due to the significant cost this would incur.

\subsection{Memorization}
\label{sec:app:api:memorization}
The Spider development set is available on GitHub, and is therefore possibly in the training set of Codex.
We believe that this does not manifest as memorization for our results however, for the following reasons.

Evaluation data on Spider's repo is formatted differently to our prompts.
Most related is the \href{https://github.com/taoyds/spider/blob/master/evaluation_examples/dev.sql}{\texttt{dev.sql} file}, which contains evaluation question-query pairs in the following format: 
\begin{verbatim}
Question 1: …
SQL: …
…
\end{verbatim}
This resembles but isn't identical to our "Question" prompt.
We prompted Codex with verbatim fragments of this file and generations failed to replicate any file contents.
Our "Question" prompt has very poor performance - hardly an indication of memorization from \texttt{dev.sql}.
Furthermore, most of Codex's performance is due to including in the prompt the schemas (see \Cref{tab:prompts}), which are not present in \texttt{dev.sql}.

As well, Codex prediction style is very different to evaluation gold queries.
Gold queries make use of a consistent table aliasing strategy (using T1, T2, etc.) which we never see with Codex (see \Cref{fig:extra_error_examples} for example comparisons).

Furthermore, in \Cref{tab:full_results} we reported performance for all models on spider-realistic \citep{Deng_2021}, a modification of the spider evaluation set that removes column name references in questions.
We observe a similar trend in performance across models as on spider (the consistent performance drop on spider-realistic is expected due to the difficulty of the updated dataset).
Memorization cannot account for the performance observed, as spider-realistic is not publicly available on GitHub.

Finally, \citet{ziegler_research_2021} studied memorization in Copilot, a derivative of the Codex models, and found that "Copilot can quote a body of code verbatim, but that it rarely does so, and when it does, it mostly quotes code that everybody quotes, and mostly at the beginning of a file".
Spider evaluation data is rare on GitHub, and we use long contexts in our prompts that significantly differ from the files on GitHub.

\subsection{Choice of Spider Evaluation Set}
We chose not to evaluate on the held-out test set of Spider, as this could not be done offline - it would instead require sending these held-out examples through the API to OpenAI, which risks inadvertently leaking them for retraining of Codex.

\onecolumn
\clearpage

\section{Additional Tables and Figures}
\label{sec:app_addition}

\begin{table*}[h!]
    \centering
    \begin{tabular}{lllccc}
        \hline
        \textbf{Engine} & \textbf{Prompt} & \textbf{VA} & \textbf{EX} & \textbf{TS} \\
        \hline
        \multicolumn{5}{l}{\textit{GPT-3}} \\
        \hline
        ada & Question & 1.2 (1.0) & 0.0 (0.0) & 0.0 (0.0) \\
        ada & Docs & 3.4 (2.2) & 0.2 (0.2) & 0.1 (0.0) \\
        ada & 1 Row & 40.1 (34.6) & 1.1 (0.6) & 0.2 (0.0) \\
        ada & Schema & 33.8 (33.9) & 2.3 (3.5) & 0.3 (0.0) \\
        \hline
        babbage & Question & 4.4 (2.0) & 1.0 (0.2) & 1.0 (0.2) \\
        babbage & Docs & 22.5 (20.3) & 1.0 (0.6) & 0.7 (0.2) \\
        babbage & 1 Row & 56.0 (49.8) & 5.1 (1.6) & 3.9 (0.0) \\
        babbage & Schema & 48.8 (44.9) & 5.7 (0.8) & 3.9 (0.0) \\
        \hline
        curie & Question & 9.0 (6.7) & 2.9 (2.4) & 2.5 (1.8) \\
        curie & Docs & 25.2 (25.0) & 7.4 (5.5) & 6.3 (3.3) \\
        curie & 1 Row & 70.6 (67.3) & 10.8 (7.3) & 7.6 (1.4) \\
        curie & Schema & 70.9 (72.2) & 12.6 (11.0) & 8.3 (4.1) \\
        \hline
        davinci & Schema & 65.0 (65.4) & 26.3 (23.2) & 21.7 (14.2) \\
        
        \hline
        \multicolumn{5}{l}{\textit{Finetuned GPT-3}} \\
        \hline
        ada & Schema & 27.5 (21.3) & 20.2 (14.0) & 19.1 (13.0) \\
        babbage & Schema & 47.2 (38.0) & 34.8 (23.6) & 31.9 (20.9) \\
        curie & Schema & 66.9 (60.2) & 51.3 (37.8) & 46.9 (32.9) \\

        \hline
        \multicolumn{5}{l}{\textit{Codex}} \\
        \hline
        cushman & Question & 11.3 (8.1) & 8.5 (3.9) & 8.3 (3.9) \\
        cushman & Docs & 83.8 (80.5) & 53.2 (45.1) & 43.5 (32.3) \\
        cushman & 1 Row & 84.7 (80.9) & 59.6 (49.2) & 48.5 (32.5) \\
        cushman & 3 Rows & 82.9 (79.1) & 60.3 (49.2) & 49.4 (33.7) \\
        cushman & 5 Rows & 83.6 (78.3) & 61.5 (49.6) & 50.4 (33.9) \\
        cushman & Schema & 88.3 (83.1) & 62.1 (49.6) & 53.1 (36.2) \\
        cushman & + 1 Row & 86.3 (85.0) & 63.7 (54.9) & 53.0 (39.6) \\
        \hline
        davinci & Question & 14.0 (8.9) & 8.3 (4.5) & 8.2 (4.1) \\
        davinci & Docs & 83.8 (87.4) & 56.8 (51.8) & 47.5 (39.0) \\
        davinci & 1 Row & 86.3 (83.5) & 60.9 (54.7) & 52.0 (41.3) \\
        davinci & 3 Rows & 85.8 (82.7) & 60.3 (53.3) & 52.2 (40.0) \\
        davinci & 5 Rows & 85.2 (80.9) & 60.5 (51.4) & 51.5 (38.4) \\
        davinci & 10 Rows & 86.0 (80.7) & 60.8 (53.3) & 51.2 (39.2) \\
        davinci & Schema & 89.8 (87.8) & 59.9 (52.2) & 50.0 (38.4) \\
        davinci & + 1 Row & 92.5 (90.7) & 64.8 (58.7) & 53.7 (41.7) \\
        davinci & + 3 Rows & 91.6 (90.6) & 67.0 (60.2) & 55.1 (42.9) \\
        davinci & + 5 Rows & 91.0 (90.2) & 65.3 (60.0) & 53.9 (42.1) \\
        davinci & + 10 Rows & 91.2 (90.6) & 63.3 (58.3) & 52.4 (40.2) \\

    \end{tabular}
    \caption{
    Performance on Spider across all evaluated models and prompts, as measured by percentage of predictions which are valid/executable SQL (VA), execution accuracy (EX), test-suite accuracy (TS).
    Main results are on the Spider development set, results in parantheses are on Spider-Realistic~\citep{Deng_2021}, a modified subset of the Spider development set with explicit references to column names removed from questions.
    }
    \label{tab:full_results}
\end{table*}

\clearpage

\begin{figure*}[]
    \centering
    \small
    \setlength\tabcolsep{4pt}
    \begin{tabularx}{\linewidth}{l X}
        14\% & \textbf{Shortcuts}\\
        \textbf{Q} & \textit{What are the template ids with template type description "Presentation".}\\
        \textbf{Gold} & \texttt{SELECT T2.template\_id FROM Ref\_template\_types AS T1 JOIN Templates AS T2 ON T1.template\_type\_code  = T2.template\_type\_code WHERE T1.template\_type\_description  =  "Presentation"}\\
        \textbf{Pred} & \texttt{SELECT Template\_ID FROM Templates WHERE Template\_Type\_Code = 'PPT';}\\
        \hline

        8\% & \textbf{SELECT Extra Columns}\\
        \textbf{NL} & \textit{Which city has the most frequent destination airport?}\\
        \textbf{Gold} & \texttt{SELECT T1.City FROM AIRPORTS AS T1 JOIN FLIGHTS AS T2 ON T1.AirportCode  =  T2.DestAirport GROUP BY T1.City ORDER BY count(*) DESC LIMIT 1}\\
        \textbf{Pred} & \texttt{SELECT City, COUNT(*) AS 'Number of Destinations' FROM airports JOIN flights ON airports.AirportCode = flights.DestAirport GROUP BY City ORDER BY COUNT(*) DESC LIMIT 1;}\\
        \hline
        
        5\% & \textbf{SELECT Convention}\\
        \textbf{NL} & \textit{Which airlines have a flight with source airport AHD?} \\
        \textbf{Gold} & \texttt{SELECT T1.Airline FROM AIRLINES AS T1 JOIN FLIGHTS AS T2 ON T1.uid  =  T2.Airline WHERE T2.SourceAirport  =  "AHD"} \\
        \textbf{Pred} & \texttt{SELECT Airline FROM flights WHERE SourceAirport = 'AHD';} \\
        \hline
        
        4\% & \textbf{GROUP BY Convention}\\
        \textbf{NL} & \textit{Which makers designed more than 3 car models? List full name and the id.}\\
        \textbf{Gold} & \texttt{SELECT T1.FullName ,  T1.Id FROM CAR\_MAKERS AS T1 JOIN MODEL\_LIST AS T2 ON T1.Id  =  T2.Maker GROUP BY T1.Id HAVING count(*)  >  3;}\\
        \textbf{Pred} & \texttt{SELECT car\_makers.FullName, car\_makers.Id FROM car\_makers JOIN model\_list ON car\_makers.Id = model\_list.Maker GROUP BY car\_makers.FullName HAVING COUNT(model\_list.Model) > 3;}\\
        \hline
        
        4\% & \textbf{Argmax}\\
        \textbf{NL} & \textit{Give the name, year of independence, and surface area of the country that has the lowest population.} \\
        \textbf{Gold} & \texttt{SELECT Name, SurfaceArea, IndepYear FROM country ORDER BY Population LIMIT 1} \\
        \textbf{Pred} & \texttt{SELECT Name, IndepYear, SurfaceArea FROM country WHERE Population = (SELECT MIN(Population) FROM country);} \\
        \hline
        
        51\% & \textbf{Other - Semantic Incorrect}\\
        \textbf{NL} & \textit{Which countries have either English or Dutch as an official language?}\\
        \textbf{Gold} & \texttt{SELECT * FROM country AS T1 JOIN countrylanguage AS T2 ON T1.Code  =  T2.CountryCode WHERE T2.Language  =  "English" AND IsOfficial  =  "T" UNION SELECT * FROM country AS T1 JOIN countrylanguage AS T2 ON T1.Code  =  T2.CountryCode WHERE T2.Language  =  "Dutch" AND IsOfficial  =  "T"}\\
        \textbf{Pred} & \texttt{SELECT * FROM countrylanguage WHERE IsOfficial = 'T' OR IsOfficial = 'F';}\\
        \hdashline
        \textbf{NL} & \textit{Show the name and the release year of the song by the youngest singer.}\\
        \textbf{Gold} & \texttt{SELECT song\_name ,  song\_release\_year FROM singer ORDER BY age LIMIT 1}\\
        \textbf{Pred} & \texttt{SELECT Name, Song\_release\_year FROM singer WHERE Age = (SELECT MIN(Age) FROM singer);}\\
        \hdashline
        \textbf{NL} & \textit{What is the maximum miles per gallon of the car with 8 cylinders or produced before 1980 ?}\\
        \textbf{Gold} & \texttt{select max(mpg) from cars\_data where cylinders  =  8 or year  <  1980}\\
        \textbf{Pred} & \texttt{SSELECT MAX(MPG) FROM cars\_data WHERE Cylinders = 8 AND Year < 1980;}\\
        \hline
        
        14\% & \textbf{Other - Ambiguous Correct}\\
        \textbf{NL} & \textit{Find the first name and age of students who have a pet.}\\
        \textbf{Gold} & \texttt{SELECT DISTINCT T1.fname ,  T1.age FROM student AS T1 JOIN has\_pet AS T2 ON T1.stuid  =  T2.stuid}\\
        \textbf{Pred} & \texttt{SELECT Fname, Age FROM Student WHERE StuID IN (SELECT StuID FROM Has\_Pet);}\\
        \hdashline
        \textbf{NL} & \textit{How many countries speak both English and Dutch?}\\
        \textbf{Gold} & \texttt{SELECT COUNT(*) FROM (SELECT T1.Name FROM country AS T1 JOIN countrylanguage AS T2 ON T1.Code  =  T2.CountryCode WHERE T2.Language  =  "English" INTERSECT SELECT T1.Name FROM country AS T1 JOIN countrylanguage AS T2 ON T1.Code  =  T2.CountryCode WHERE T2.Language  =  "Dutch")}\\
        \textbf{Pred} & \texttt{SELECT COUNT(*) FROM countrylanguage WHERE Language = 'Dutch' AND CountryCode IN (SELECT CountryCode FROM countrylanguage WHERE Language = 'English');}\\
        \hdashline
        \textbf{NL} & \textit{What are the Asian countries which have a population larger than that of any country in Africa?}\\
        \textbf{Gold} & \texttt{SELECT Name FROM country WHERE Continent  =  "Asia"  AND population  >  (SELECT min(population) FROM country WHERE Continent  =  "Africa")}\\
        \textbf{Pred} & \texttt{SELECT Name FROM country WHERE Continent = 'Asia' AND Population > (SELECT MAX(Population) FROM country WHERE Continent = 'Africa');}\\
        \hline
    \end{tabularx}
    \caption{
    Additional examples of error types, as made by davinci-codex model with Create Table + Select 3 prompt.
    NL stands for natural language question.
    Percentage indicates the percentage of errors which are of the given type.
    }
    \label{fig:extra_error_examples}
\end{figure*}

\clearpage

\begin{figure*}[]
    \centering
    \small
    \setlength\tabcolsep{4pt}
    \begin{tabularx}{\linewidth}{l X}
         & \textbf{10-shot examples} \\ 
         & { \texttt{what states does the missouri river run through         SELECT RIVERalias0.TRAVERSE FROM RIVER AS RIVERalias0 WHERE RIVERalias0.RIVER\_NAME = "missouri" ;         -- what is the size of texas         SELECT STATEalias0.AREA FROM STATE AS STATEalias0 WHERE STATEalias0.STATE\_NAME = "texas" ;                  -- what are the major cities in texas         SELECT CITYalias0.CITY\_NAME FROM CITY AS CITYalias0 WHERE CITYalias0.POPULATION > 150000 AND CITYalias0.STATE\_NAME = "texas" ;                  -- what is the capital of pennsylvania         SELECT STATEalias0.CAPITAL FROM STATE AS STATEalias0 WHERE STATEalias0.STATE\_NAME = "pennsylvania" ;                  -- what is the biggest city in nebraska         SELECT CITYalias0.CITY\_NAME FROM CITY AS CITYalias0 WHERE CITYalias0.POPULATION = ( SELECT MAX( CITYalias1.POPULATION ) FROM CITY AS CITYalias1 WHERE CITYalias1.STATE\_NAME = "nebraska" ) AND CITYalias0.STATE\_NAME = "nebraska" ;                  -- what is the population of austin         SELECT CITYalias0.POPULATION FROM CITY AS CITYalias0 WHERE CITYalias0.CITY\_NAME = "austin" ;                  -- which state is kalamazoo in         SELECT CITYalias0.STATE\_NAME FROM CITY AS CITYalias0 WHERE CITYalias0.CITY\_NAME = "kalamazoo" ;                  -- name all the rivers in colorado         SELECT RIVERalias0.RIVER\_NAME FROM RIVER AS RIVERalias0 WHERE RIVERalias0.TRAVERSE = "colorado" ;                  -- what states border missouri         SELECT BORDER\_INFOalias0.BORDER FROM BORDER\_INFO AS BORDER\_INFOalias0 WHERE BORDER\_INFOalias0.STATE\_NAME = "missouri" ;                  -- how many people live in new mexico         SELECT STATEalias0.POPULATION FROM STATE AS STATEalias0 WHERE STATEalias0.STATE\_NAME = "new mexico" ;}} \\

         & \textbf{Very similar query in the few-shot prompt fixes the example}\\
        \textbf{Question} & \textit{which states border iowa}\\
        \textbf{Gold} & \texttt{SELECT BORDER\_INFOalias0.BORDER FROM BORDER\_INFO AS BORDER\_INFOalias0 WHERE BORDER\_INFOalias0.STATE\_NAME = "iowa" ;}\\
        
        \textbf{0-shot pred} & \texttt{SELECT state\_name FROM border WHERE border = 'iowa'}\\
        \textbf{10-shot pred} & \texttt{SELECT BORDER\_INFOalias0.BORDER FROM BORDER\_INFO AS BORDER\_INFOalias0 WHERE BORDER\_INFOalias0.STATE\_NAME = "iowa"}\\
        \hline

         & \textbf{Argmax convetion fixed}\\
        \textbf{Question} & \textit{what state has the smallest population}\\
        \textbf{Gold} & \texttt{SELECT STATEalias0.STATE\_NAME FROM STATE AS STATEalias0 WHERE STATEalias0.POPULATION = (SELECT MIN(STATEalias1.POPULATION) FROM STATE AS STATEalias1) ;}\\
        \textbf{0-shot pred} & \texttt{SELECT state\_name FROM state ORDER BY population LIMIT 1}\\
        \textbf{10-shot pred} & \texttt{SELECT STATEalias0.STATE\_NAME FROM STATE AS STATEalias0 WHERE STATEalias0.POPULATION = (SELECT MIN(STATEalias1.POPULATION) FROM STATE AS STATEalias1)}\\
        \hline
        
         & \textbf{SELECT extra columns fixed}\\
        \textbf{Question} & \textit{what is the population of the state with the largest area}\\
        \textbf{Gold} & \texttt{SELECT STATEalias0.POPULATION FROM STATE AS STATEalias0 WHERE STATEalias0.AREA = (SELECT MAX(STATEalias1.AREA) FROM STATE AS STATEalias1) ;}\\
        \textbf{0-shot pred} & \texttt{SELECT state\_name, population FROM state WHERE area = (SELECT MAX(area) FROM state)}\\
        \textbf{10-shot pred} & \texttt{SELECT STATEalias0.POPULATION FROM STATE AS STATEalias0 WHERE STATEalias0.AREA = (SELECT MAX(STATEalias1.AREA) FROM STATE AS STATEalias1)}\\
        
        \hline
    \end{tabularx}
    \caption{
    Cherry-picked examples of Codex improvements from 0-shot to 10-shot text-to-SQL on GeoQuery validation set.
    The style of the generated SQL changes a lot and is much closer to that of the gold SQL when few-shot examples are in the prompt.
    The few-shot examples were also useful to adapt the generated SQL to the conventions of the dataset, like the way argmax is done, or the selected columns.
    }
    \label{fig:extra_error_examples_fewshot}
\end{figure*}

\clearpage

\section{Example Prompts}
\label{sec:appendix_prompts}

\begin{figure*}[h]
    \centering
    \small
\begin{lstlisting}[breaklines]
What is Kyle's id? | network_1 | highschooler : id, name ( Kyle ), grade | friend : student_id, friend_id | likes : student_id, liked_id
\end{lstlisting}
    \caption{Example input for baseline T5 models.}
    \label{fig:prompt_question}
\end{figure*}

\begin{figure*}[h]
    \centering
    \small
\begin{lstlisting}[breaklines]
-- Using valid SQLite, answer the following questions.

-- What is Kyle's id?
SELECT
\end{lstlisting}
    \caption{Example prompt for \textbf{Question}.}
    \label{fig:prompt_question}
\end{figure*}

\begin{figure*}[h]
    \centering
    \small
\begin{lstlisting}[breaklines]
### SQLite SQL tables, with their properties:
#
# Highschooler(ID, name, grade)
# Friend(student_id, friend_id)
# Likes(student_id, liked_id)
#
### What is Kyle's id?
SELECT
\end{lstlisting}
    \caption{Example prompt for \textbf{API Docs}.}
    \label{fig:prompt_documentation}
\end{figure*}

\begin{figure*}[h]
    \centering
    \small
\begin{lstlisting}[breaklines]
/*
3 example rows from table Highschooler:
SELECT * FROM Highschooler LIMIT 3;
Table: Highschooler
  ID    name  grade
1510  Jordan      9
1689 Gabriel      9
1381 Tiffany      9
*/

/*
3 example rows from table Friend:
SELECT * FROM Friend LIMIT 3;
Table: Friend
 student_id  friend_id
       1510       1381
       1510       1689
       1689       1709
*/

/*
3 example rows from table Likes:
SELECT * FROM Likes LIMIT 3;
Table: Likes
 student_id  liked_id
       1689      1709
       1709      1689
       1782      1709
*/


-- Using valid SQLite, answer the following questions for the tables provided above.

-- What is Kyle's id?
SELECT
\end{lstlisting}
    \caption{Example prompt for \textbf{Select 3}.}
    \label{fig:prompt_rows}
\end{figure*}

\begin{figure*}
    \centering
    \small
\begin{lstlisting}[breaklines]
CREATE TABLE Highschooler(
        ID int primary key,
        name text,
        grade int)

CREATE TABLE Friend(
        student_id int,
        friend_id int,
        primary key (student_id,friend_id),
        foreign key(student_id) references Highschooler(ID),
        foreign key (friend_id) references Highschooler(ID)
)

CREATE TABLE Likes(
        student_id int,
        liked_id int,
        primary key (student_id, liked_id),
        foreign key (liked_id) references Highschooler(ID),
        foreign key (student_id) references Highschooler(ID)
)


-- Using valid SQLite, answer the following questions for the tables provided above.

-- What is Kyle's id?
SELECT
\end{lstlisting}
    \caption{Example prompt for \textbf{Create Table}.}
    \label{fig:prompt_schema}
\end{figure*}

\begin{figure*}
    \centering
    \small
\begin{lstlisting}[breaklines]
CREATE TABLE Highschooler(
        ID int primary key,
        name text,
        grade int)
/*
3 example rows:
SELECT * FROM Highschooler LIMIT 3;
  ID    name  grade
1510  Jordan      9
1689 Gabriel      9
1381 Tiffany      9
*/

CREATE TABLE Friend(
        student_id int,
        friend_id int,
        primary key (student_id,friend_id),
        foreign key(student_id) references Highschooler(ID),
        foreign key (friend_id) references Highschooler(ID)
)
/*
3 example rows:
SELECT * FROM Friend LIMIT 3;
 student_id  friend_id
       1510       1381
       1510       1689
       1689       1709
*/

CREATE TABLE Likes(
        student_id int,
        liked_id int,
        primary key (student_id, liked_id),
        foreign key (liked_id) references Highschooler(ID),
        foreign key (student_id) references Highschooler(ID)
)
/*
3 example rows:
SELECT * FROM Likes LIMIT 3;
 student_id  liked_id
       1689      1709
       1709      1689
       1782      1709
*/


-- Using valid SQLite, answer the following questions for the tables provided above.

-- What is Kyle's id?
SELECT
\end{lstlisting}
    \caption{Example prompt for \textbf{Create Table + Select 3}.}
    \label{fig:prompt_schema_and_examples}
\end{figure*}

\begin{figure*}
    \centering
    \tiny
    \begin{lstlisting}[breaklines]
CREATE TABLE "border_info" ("state_name" text, "border" text)
/*
state_name    border
   alabama tennessee
   alabama   georgia
   alabama   florida
*/

CREATE TABLE "city" ("city_name" text, "population" int DEFAULT NULL, "country_name" varchar(3) NOT NULL DEFAULT '', "state_name" text)
/*
 city_name  population country_name state_name
birmingham      284413          usa    alabama
    mobile      200452          usa    alabama
montgomery      177857          usa    alabama
*/

CREATE TABLE "highlow" ("state_name" text, "highest_elevation" text, "lowest_point" text, "highest_point" text, "lowest_elevation" text)
/*
state_name highest_elevation   lowest_point   highest_point lowest_elevation
   alabama               734 gulf of mexico cheaha mountain                0
    alaska              6194  pacific ocean  mount mckinley                0
   arizona              3851 colorado river  humphreys peak               21
*/

CREATE TABLE "lake" ("lake_name" text, "area" double DEFAULT NULL, "country_name" varchar(3) NOT NULL DEFAULT '', "state_name" text)
/*
lake_name   area country_name state_name
  iliamna 2675.0          usa     alaska
 becharof 1186.0          usa     alaska
teshekpuk  816.0          usa     alaska
*/

CREATE TABLE "mountain" ("mountain_name" text, "mountain_altitude" int DEFAULT NULL, "country_name" varchar(3) NOT NULL DEFAULT '', "state_name" text)
/*
mountain_name  mountain_altitude country_name state_name
     mckinley               6194          usa     alaska
    st. elias               5489          usa     alaska
      foraker               5304          usa     alaska
*/

CREATE TABLE "river" ("river_name" text, "length" int DEFAULT NULL, "country_name" varchar(3) NOT NULL DEFAULT '', "traverse" text)
/*
 river_name  length country_name  traverse
mississippi    3778          usa minnesota
mississippi    3778          usa wisconsin
mississippi    3778          usa      iowa
*/

CREATE TABLE "state" ("state_name" text, "population" int DEFAULT NULL, "area" double DEFAULT NULL, "country_name" varchar(3) NOT NULL DEFAULT '', "capital" text, "density" double DEFAULT NULL)
/*
state_name  population     area country_name    capital   density
   alabama     3894000  51700.0          usa montgomery 75.319149
    alaska      401800 591000.0          usa     juneau  0.679865
   arizona     2718000 114000.0          usa    phoenix 23.842105
*/

-- Using valid SQLite, answer the following questions for the tables provided above.
-- what is the population of austin
SELECT CITYalias0.POPULATION FROM CITY AS CITYalias0 WHERE CITYalias0.CITY_NAME = "austin" ;

-- which state is kalamazoo in
SELECT CITYalias0.STATE_NAME FROM CITY AS CITYalias0 WHERE CITYalias0.CITY_NAME = "kalamazoo" ;

-- name all the rivers in colorado
SELECT RIVERalias0.RIVER_NAME FROM RIVER AS RIVERalias0 WHERE RIVERalias0.TRAVERSE = "colorado" ;

-- how many people live in new mexico
SELECT STATEalias0.POPULATION FROM STATE AS STATEalias0 WHERE STATEalias0.STATE_NAME = "new mexico" ;

-- what states border missouri
SELECT BORDER_INFOalias0.BORDER FROM BORDER_INFO AS BORDER_INFOalias0 WHERE BORDER_INFOalias0.STATE_NAME = "missouri" ;

-- what is the biggest city in arizona
SELECT
    \end{lstlisting}
    \caption{
    Example prompt for \textbf{5-shot}.
    It starts with the schema and 3 rows per database (exactly as in ~\Cref{fig:prompt_schema_and_examples}), followed by 5 few-shot examples, and finally the target question.}
    \label{fig:prompt_fewshot}
\end{figure*}

\end{document}